\documentclass[wcp]{jmlr}

\usepackage{longtable}
\usepackage{booktabs}
\usepackage{algorithm}
\usepackage{bbm}

\makeatletter
\let\Ginclude@graphics\@org@Ginclude@graphics 
\makeatother
\DeclareMathOperator{\var}{var}

\jmlrvolume{189}
\editors{Emtiyaz Khan and Mehmet G\"{o}nen}
\jmlryear{2022}
\jmlrworkshop{ACML 2022}

\title[Robust Direct Learning for Causal Data Fusion]{Robust Direct Learning for Causal Data Fusion}

 \author{\Name{Xinyu Li}$^{1, 2}$ \Email{xinyu.li@pku.edu.cn}\\
 \Name{Yilin Li}$^{1}$ \Email{yilinli@pku.edu.cn}\\
 \Name{Qing Cui}$^{2}$ \Email{cuiqing.cq@antgroup.com}\\
 \Name{Longfei Li}$^{2}$ \Email{longyao.llf@antgroup.com}\\
 \Name{Jun Zhou}$^{2}$\thanks{Corresponding author.} \Email{jun.zhoujun@antfin.com}\\
 \addr $^{1}$School of Mathematical Sciences, Peking University, Beijing, China\\
 \addr $^{2}$Ant Group, Hangzhou, China}

\begin{document}

\maketitle

\begin{abstract}

In the era of big data, the explosive growth  of multi-source heterogeneous data offers many exciting challenges and opportunities for improving  the inference of  conditional average treatment effects.
In this paper, we investigate homogeneous and heterogeneous causal data fusion problems under a general setting that allows for the presence of source-specific covariates. 
We provide a direct learning framework for integrating multi-source data that separates the treatment effect from other nuisance functions, and achieves double robustness against certain misspecification.
To improve estimation precision and stability, we propose a causal information-aware weighting function motivated by theoretical insights from the semiparametric efficiency theory; it assigns larger weights to samples containing more causal information with  high interpretability.
We introduce a two-step algorithm, the weighted multi-source direct learner, 
based on constructing a pseudo-outcome and regressing it on covariates under a weighted least square criterion; it  offers us a powerful tool for causal data fusion, enjoying the advantages of  easy implementation, double robustness and model flexibility. 
In simulation studies, we  demonstrate the effectiveness of  our proposed methods in both homogeneous and heterogeneous causal data fusion scenarios.

\end{abstract}

\section{Introduction}

To understand the causal mechanism, a classic parameter of interest is the conditional average treatment effects (CATE), also known as heterogeneous treatment effect \citep{abrevaya2015estimating,athey2016recursive, kunzel2019metalearners}, defined by the difference in outcome means between the two treatment groups conditional on a set of background attributes.
Learning CATE is one of the fundamental problems in experimental sciences,  observational studies, and electric commerce, such as the average effectiveness of medical treatment on patients \citep{obermeyer2016predicting} and benefits of advertising on consumers \citep{bottou2013counterfactual}. A vast number of methods have been proposed to estimate the CATE based on flexible machine learning methods, including tree-based methods \citep{athey2016recursive, tang2022debiased}, random forests \citep{wager2018estimation}, boosting \citep{powers2018some, nie2021quasi},  neural networks \citep{johansson2016learning,louizos2017causal,shi2019adapting}, and meta-learners with any supervised learning method \citep{kunzel2019metalearners,nie2021quasi}.

As pointed out by \citet{kallus2022robust}, learning the conditional outcome means on two treatment arms separately and taking the difference to obtain the CATE may suffer from error accumulation, especially when the CATE function has a simpler and sparser form. This inspires us to learn the CATE function directly by modeling it as a whole.
\citet{qi2018d, qi2020multi} proposed a one-step method (D-learning) to directly learn  the optimal individual treatment rule which is closely related to the CATE. \citet{meng2020doubly} further generalized  D-learning by replacing the outcome with the residual of some main effect function to achieve double robustness. The double robustness property is well studied in causal inference, meaning that the estimation is consistent if either the propensity score or the conditional outcome mean model is correct but not necessarily both, see \citet{bang2005doubly, zhang2012robust}. 

Although various methods have been proposed for estimating the CATE on a single dataset, the practical performance might be poor due to the limited sample size.
A natural idea is to combine other similar datasets and improve the precision of the estimating procedures.
Integrating and leveraging data from multiple sources have received wide attention in recent years. This problem is typically known as \textit{causal data fusion}. Some notable advances focus on the average treatment effect (ATE) \citep{fan2014identifying,bareinboim2016causal,colnet2020causal,li2021improving}, causal discovery \citep{claassen2010causal,zhang2017causal} across multiple data sources. Recently, a tree-based approach for estimating the CATE is proposed when the individual-level data cannot be pooled \citep{tan2021tree}. 

To improve efficiency, we propose a novel approach for estimating the CATE on heterogeneous data sources by generalizing the approach from \cite{meng2020doubly}. We have the following four concrete contributions: (\romannumeral1) We formulate and investigate the homogeneous and heterogeneous causal data fusions  under general
settings that allow for the presence of source-specific covariates. (\romannumeral2) We present a multi-source direct learning framework, and propose a direct, model-flexible and doubly robust algorithm for causal data fusion. (\romannumeral3) We propose a causal information-aware weighting  function based on semiparametric efficiency theory to improve  efficiency. (\romannumeral4) We demonstrate the performance of the proposed methods and show the improvement compared with other methods.

\section{Preliminaries and notations}\label{sec:2}

\subsection{Heterogeneous Treatment Effect}\label{sec:2.1}

As is customary, we use capital letters for random variables and lowercase letters for realized values; in particular,  let $Y$ denote the outcome of interest, $A\in \{1,-1\}$ denote a binary
treatment indicator, $X\in \mathcal{X}$ denote a vector of features, and $Y(a)$ denote the potential outcome that would be observed when the treatment $A$ had been set
to $a \in \{1, -1\}$ \citep{rubin1974estimating, imbens2015causal}. We maintain the classic
stable unit treatment value assumption (SUTVA) that no interference between units and no hidden variations
of treatments occur, and assume that the observed outcome is a realization
of the potential outcome under the intervention actually received, i.e., $Y=\sum_{a}\mathbbm{1}(A=a)Y(a)$, where $\mathbbm{1}(\cdot)$ denotes the indicator function.

The CATE is defined as the mean difference in potential outcomes between the treated and control groups for individuals with feature $X$, that is, $\tau(X) = \mu_1(X)-\mu_{-1}(X)$, where $\mu_a(X)=E[Y(a)\mid X]$ for $a\in \{1,-1\}$; it captures the heterogeneity of average treatment effect  across subpopulations defined by the values of the features.


We consider the following general model  for potential outcomes,
$Y(a)=\mu_a(X)+\tilde{\varepsilon}_a$,
where $\tilde{\varepsilon}_a$ is a random error satisfying  $E(\tilde{\varepsilon}_a\mid X)=0$ and $\var(\tilde\varepsilon_a)<\infty$ for $a\in\{1,-1\}$.
Then because $Y=\sum_{a}\mathbbm{1}(A=a)Y(a)$,  the observed outcome can be naturally modeled by
\begin{align}
    Y&= \frac{\mu_1(X)+\mu_{-1}(X)}{2}+A\frac{\mu_1(X)-\mu_{-1}(X)}{2}+\varepsilon \nonumber\\
    &\triangleq m(X)+A\delta(X)+\varepsilon,\label{eq1}
\end{align}
where $E(\varepsilon\mid X,A)=0$, $\var(\varepsilon)<\infty$.
Hereafter we refer to $m(X)$ as the main effect function and $\delta(X)$ as the treatment effect function.
Note that $\tau(X)=2\delta(X)$, and thus estimating the CATE is equivalent to  estimating  the treatment effect function $\delta(X)$.

\subsection{Multiple Heterogeneous Data Sources}

Suppose that data  are available from  $K$ mutually independent  data sources; in each of them, an individual either receives the experimental or the control intervention.
When data across different sources are samples from the same global population, we refer to the data sources as \textit{homogeneous}, and otherwise as \textit{heterogeneous};  here we consider the latter, a more challenging case that  allows for differences in the distribution of baseline features across sources.
We introduce $S$ as the source indicator which takes values in $\mathcal{S}=\{1,\cdots,K\}$, and denote the dataset  observed from the $s$-th data source as $\mathcal{O}_s$.

We consider the problem where observed covariates differ across multiple datasets; such scenarios are common in practice \citep[e.g.,][]{jia2006identifiability,li2022causal}.
In particular, we denote the covariates of interest that are  shared by all datasets as $X$, and the covariates of the $s$-th dataset other than $X$ as $Z_s$. 
For each data source $s$,   the observed dataset $\mathcal{O}_s=\{(Y_i, A_i, X_i, Z_{s,i}),i=1,\cdots,n_s \}$, in which the samples  are independent and identically distributed according to $f(Y,A,X, Z_s \mid S=s)$, where $f(\cdot)$ denotes the density function.
If $Z_s=\varnothing$ for each $s\in \mathcal{S}$, then it degenerates to the classic setting where all covariates are observed across data sources \citep{colnet2020causal}.
In the following, if not otherwise specified, we denote  $E(\cdot)$ the expectation with respect to the joint distribution of observed data $\prod_{s=1}^{K}\left[f(Y,A,X, Z_s,S=s)\right]^{\mathbbm{1}(S=s)}$, and $\hat E(\cdot)$ the empirical expectation.

We are interested in drawing inference about the causal effects  conditional on the  commonly shared covariates $X$, and the causal quantity of interest is the CATE function  $\tau_s(X)=E\{Y(1)-Y(-1)\mid X,S=s\}=E[E\{Y(1)-Y(-1)\mid X,Z_s,S=s\}\mid X,S=s]$ of each data source. To identify the CATE, we impose the following regularity assumptions \citep{imbens2015causal}.
\begin{assumption}[Ignorability]\label{assp:1}
$Y(a)\perp A\mid X,Z_s,S=s$ for each $s$.
\end{assumption}
\begin{assumption}[Positivity]\label{assp:2}
There exist a constant $c>0$ such that  $P(A=a \mid X,Z_s,S=s)>c$ almost surely for each $a$ and $s$.
\end{assumption}

Assumption \ref{assp:1} excludes the unmeasured confounders between $A$ and $Y(a)$ within each data source, which is plausible as in many cases the assignment of interventions within each data source can be characterized by observed features.   Assumption \ref{assp:2} requires that populations of  the treated and control group have some overlap.
Within each data source, under Assumptions \ref{assp:1}--\ref{assp:2}, the CATE function $\tau_s(X)$ is identified, and
a straightforward estimating method is to utilize data from the $s$-th source solely.
However, it is appealing to integrate all data from different sources to enhance efficiency, especially when the sample size of each source is relatively small.

\section{Direct Learning for Causal Data Fusion}\label{sec:3}

\subsection{Causal Data Fusion}

We define the main and the treatment effect function of the source $s$ as
\begin{equation*}
\begin{aligned}
m_s(X,Z_s)=\frac{\mu_{1s}(X,Z_s)+\mu_{-1s}(X,Z_s)}{2} \text{ and }
\delta_s(X,Z_s)=\frac{\mu_{1s}(X,Z_s)-\mu_{-1s}(X,Z_s)}{2},
\end{aligned}
\end{equation*} 
respectively, where  $\mu_{as}(X,Z_s)=E\{Y(a)\mid X,Z_s,S=s\}$.   Following the same derivation as for \equationref{eq1}, in the scenario of  multiple data sources, one can  naturally model the observed outcome by
\begin{equation}\label{3.1}
Y=m_S(X,Z_S)+A\delta_S(X,Z_S)+\varepsilon_S,
\end{equation}
where $E(\varepsilon_s\mid X,Z_s,A,S=s)=0$ and $\var(\varepsilon_s)<\infty$ for $s=1,\cdots,K$.
When there is no confusion, we let $\delta_s(X)=E\{\delta_s(X,Z_s)\mid X,S=s\}$ and also refer to it  as the treatment effect function. By definition it follows that $\delta_s(X)=\tau_s(X)/2$, hence
estimating the CATE $\tau_s(X)$ is  equivalent to estimating the treatment effect function $\delta_s(X)$.

Without loss of generality, we set the first data source as the target one, i.e.,  its samples are drawn from the population  over which we are interested in inferring causal effects. For example, the first dataset $\mathcal{O}_1$  is collected from a randomized controlled trial (RCT) among the overall population  of scientific interest, whereas the second dataset $\mathcal{O}_2$ is obtained from an observational study over a specific sub-population.

We classify causal data fusion tasks under  multiple heterogeneous data sources   into two categories:
\begin{itemize}
\item  homogeneous causal data fusion, where the conditional average treatment effect across different data sources are homogeneous, see Assumption \ref{assp:3} described below;
\item heterogeneous causal data fusion, where at least one of the data sources has a different conditional mean treatment effect than the others, and we will discuss it later.
\end{itemize}

\begin{assumption}[Homogeneity of CATE]\label{assp:3}
$\delta_1(X)=\cdots=\delta_K(X)\triangleq \delta(X)$ almost surely.
\end{assumption}

Assumption \ref{assp:3} reveals that for units with the same value of covariates $X$, the expected treatment effect remains the same regardless of the data source.
It is plausible in many real cases, and weaker than the mean exchangeability or distribution exchangeability assumptions commonly adopted in causal inference literature when dealing with data fusion \citep[e.g.,][]{rudolph2017robust,buchanan2018generalizing,dahabreh2020towards,li2021improving}.

\subsection{Direct Learning for Homogeneous Causal Data Fusion}

We propose a direct learning approach  to estimate the treatment
effect function $\delta(X)$ without modeling each $\delta(X,Z_s)$. Unlike traditional methods that model  CATE through modeling  both $\mu_{1s}$ and $\mu_{-1s}$,  the direct learner separates $m_s$ and $\delta$ in the estimation procedure, allowing the use of a flexible model
for  the quantity of interest CATE; for example, one can characterize CATE with a tree model with better interpretation, but fit the nuisance functions with a more complex and powerful model. Further, we also show the robustness of the direct learner against the failure of   nuisance estimators.

In addition to main effect functions, the direct learners may also need to model  a few other  nuisance functions.
We denote the treatment propensity scores $P(A=a\mid X,Z_s,S=s)$ by $p_{a\mid s}(X,Z_s)$, and selection propensity scores $P(S=s\mid X)$ by $\pi_{s}(X)$. For simplicity, when there causes no confusion, we may omit the random variables, e.g., $p_{a\mid s}$ for $p_{a\mid s}(X,Z_s)$. Note that the notations  $P_{A\mid s}=\sum_a \mathrm{I}(A=a)P_{a\mid s}$ and $P_{A\mid S}=\sum_s\sum_a \mathrm{I}(S=s,A=a)P_{a\mid s}$.

We refer to  working propensity score models,  say $\widetilde{p}_{a\mid s}(X,Z_s)$, as satisfying the \textit{partial balance} property with respect to $\delta_s(x,z_s)$, if
$$ E\{e_A(X,Z_s)\delta(X,Z_s)\mid X,S=s\}=E\{\delta(X,Z_s)\mid X,S=s\},$$
where $e_A(X,Z_s)=\widetilde{p}_{A\mid s}^{-1}(X,Z_s)/E\{\widetilde{p}_{A\mid s}^{-1}(X,Z_s)\mid X,S=s\}$, a standardized inverse probability weighting term.
We call this property \textit{partial balance} because it characterizes the
ability of the working propensity scores to adjust for the imbalance of $Z_s$ between the treated and control groups   given $X$ along the direction of $\delta(X,Z_s)$.
It is weaker than requiring the correctness of working propensity scores, because that   calls for the ability to   fully adjust for the imbalance of all covariates including both $X$ and $Z_s$. In order to train propensity score models that satisfy such partial balance property, one can first obtain an initial $\hat\delta(X,Z_s)$ using T-learner \citep{kunzel2019metalearners} and then add the term $\hat{E}[g(X)\{P_{A\mid s}^{-1}-1\}\hat\delta\mid S=s]^2$ to the loss function, where $g(X)$ is a vector of user-specified functions. The partial balance property is naturally satisfied in many cases, for example,
(a) the working propensity scores are correct, i.e.,
$\widetilde{p}_{a\mid s}(X,Z_s)={p}_{a\mid s}(X,Z_s)$;
(b) there are no source-specific covariates, i.e., $Z_s=\varnothing$; (c) only the covariates $X$ lead to the heterogeneity of causal effects, i.e., $\delta(X,Z_s)=\delta(X)$. Case (c) holds especially when many variables predict potential outcomes but only a few has a strong modulate effect \citep{kallus2022robust}.

The following result establishes the double robustness property of the direct learning approach for homogeneous causal data fusion, which is a generalization of the result in \cite{meng2020doubly}.
It  offers us two chances to obtain a correct estimate of the CATE, thus more robust than methods that only rely on a single nuisance model. 

\begin{theorem}\label{thm:1}
Suppose that Assumptions \ref{assp:1}--\ref{assp:3} hold and $\{w_s(X)\}_{s=1}^{K}$ are arbitrary positive integrable functions. Then
\begin{equation}\label{3.2}
\delta \in \mathop{\arg\min}_{l \in\{\mathcal{X} \rightarrow \mathbb{R}\}}{E}\left[\frac{w_{S}({X})}{\widetilde{p}_{A\mid S}({X},Z_S)} \left\{\frac{Y-\widetilde{m}_S({X},Z_S)}{A}- l({X})\right\}^{2}\right],
\end{equation}
if for each data source $s$, either one of the following conditions holds:

1. the working propensity scores $\widetilde{p}_{a\mid s}({X},Z_s)={p}_{a\mid s}({X},Z_s)$, or

2. the working  main effect functions $\widetilde{m}_s({X},Z_s)={m}_s({X},Z_s)$,  and the working  propensity scores $\widetilde{p}_{a\mid s}({X},Z_s)$ satisfies the partial balance property with respect to $\delta(X,Z_s)$.
\end{theorem}

\theoremref{thm:1} suggests that we can consistently estimate $\delta(X)$ through the empirical version of \equationref{3.2}
if the limiting functions of working nuisance models satisfy either one of 
the above conditions.  
This method avoids modeling the treatment effect function $\delta(X,Z_s)$ on each dataset. Instead, we integrate all the datasets to solve directly for the objective  treatment effect function of $X$.
At the same time, it allows us to separate the treatment effect from other variation independent nuisance functions, and achieve double robustness against misspecification of them.

\begin{remark}
In many cases, such as RCTs, the propensity scores are known and  often the treatment assignment mechanism only depends on the commonly shared covariates $X$.
With such prior knowledge, we simply need to fit  propensity scores on $X$, and under the same condition as \theoremref{thm:1},
\begin{equation*}
\delta \in \mathop{\arg\min}_{l \in\{\mathcal{X} \rightarrow \mathbb{R}\}} {E}\left[\frac{w_{S}({X})}{\widetilde{p}_{A\mid S}({X})} \{Y-\widetilde{m}_S({X},Z_S)-A l({X})\}^{2}\right],
\end{equation*}
if for each  $s$, either
$\widetilde{p}_{a\mid s}({X})={p}_{a\mid s}({X})$ or $\widetilde{m}_s({X},Z_s)=m_s({X},Z_s)$ holds.
\end{remark}

\subsection{Causal Information-aware Weighting Function} \label{sec:3.4}
As shown in \theoremref{thm:1}, the weight function $w_s(X)$ plays an important role in the direct learning; the choice of $w_s(X)$  affects the efficiency of estimators, but not the consistency. Users may choose a constant weight, but ideal weights should be interpretable as well as beneficial for improving estimation precision and stability.
Towards this end, we propose a weighting function motivated by the semiparametric efficiency bound \citep[also referred to as semiparametric Cramér-Rao bound, see][]{newey1990semiparametric, bickel1993efficient, tsiatis2006semiparametric,kennedy2016semiparametric}, which characterizes the amount of information contained in the observed data  for inferring target parameters.

To illustrate, suppose all datasets share the same covariates and consider the case where  covariates are discrete-valued for simplicity.
At a fixed $X=\vec x$, in the statistical model $\set{F}_{s,\vec x}=\{f(Y, A, X=\vec x\mid S=s): \text{Assumptions \ref{assp:1}--\ref{assp:2} holds with no other restrictions}\}$,
the amount of information for estimating CATE carried  in each observation of the $s$-th dataset can be measured by the semiparametric  efficiency  bound of $\tau_s(\vec x)$ . We  refer to the bound that corresponds to $s$ and $\vec x$ as $\mathcal{B}_{s,\vec x}$;  the asymptotic variance of any regular and asymptotic linear  estimator of $\tau_s(\vec x)$ based solely on $\mathcal{O}_{s}$ can be no smaller than this bound. 
Then  the relative information between  $\mathcal{O}_{s_1}$ and $\mathcal{O}_{s_2}$ with respect to $\vec x$ is
\begin{equation*}
\mathcal{B}_{s_1,\vec x}/\mathcal{B}_{s_2,\vec x}=\left\{\frac{V_{1\mid s_2}(\vec x)}{p_{1\mid s_2}(\vec x)}+\frac{V_{-1\mid s_2}(\vec x)}{p_{-1\mid s_2}(\vec x)}\right\}^{-1}\left\{\frac{V_{1\mid s_1}(\vec x)}{p_{1\mid s_1}(\vec x)}+\frac{V_{-1\mid s_1}(\vec x)}{p_{-1\mid s_1}(\vec x)}\right\},
\end{equation*}
where $s_1,s_2\in \mathcal{S}$,    
$V_{a\mid s}(X)= \var(Y_a\mid X,S=s)=\var(Y\mid X,A=a,S=s)$ and $p_{a\mid s}(X)=P(A=a\mid X,S=s)$ for $a\in\{1,-1\}$.


In light of this, we propose an information-aware weighting function as
\begin{align}\label{eq3}
w_s(X)&\propto \frac{P(S=s)}{
P(S=1)}\frac{\pi_1(X)}{\pi_s(X)} \left\{\frac{V_{1\mid s}(X)}{{p}_{1\mid s}({X})}+\frac{V_{-1\mid s}(X)}{{p}_{-1\mid s}({X})}  \right\}^{-1}\\
&= \frac{f(X\mid S=1)}{f(X\mid S=s)} \left\{\frac{V_{1\mid s}(X)}{{p}_{1\mid s}({X})}+\frac{V_{-1\mid s}(X)}{{p}_{-1\mid s}({X})}  \right\}^{-1}.\nonumber
\end{align}

\begin{figure}[h]
\floatconts
  {fig:weights}
  {\caption{Illustration of the  weighting function. Suppose that $V_{a\mid s}(\vec x)$ is a constant, and $f(X\mid S=s)$ is shown in (a) and (b).
   The blue and yellow areas indicate the proportions of the treated and control units at each value of $X$, respectively. (c), (d) and (e) show the variation of $I_s(X)$, $R_s(X)$ and $w_s(X)=I_s(X)R_s(X)$ with $X$. The solid and dashed lines stand for the first and  second sources, respectively. Here $I_s(X)$ characterizes the imbalance of treatment, while $R_s(X)$ characterizes the imbalance of populations between two sources. 
   When $X$ is close to one, $R_2(X)>R_1(X)$ which enables the second source population to match the first one, and $I_2(X)>I_1(X)$ because the first source suffers from a severe treatment imbalance; in this case, the weighting function $w_s(X)$ assigns more weight to the samples from the second source. 
    }}
  {%
    \subfigure[$f(X\mid S=1)$]{\label{fig:den1}
      \includegraphics[width=0.3\linewidth]{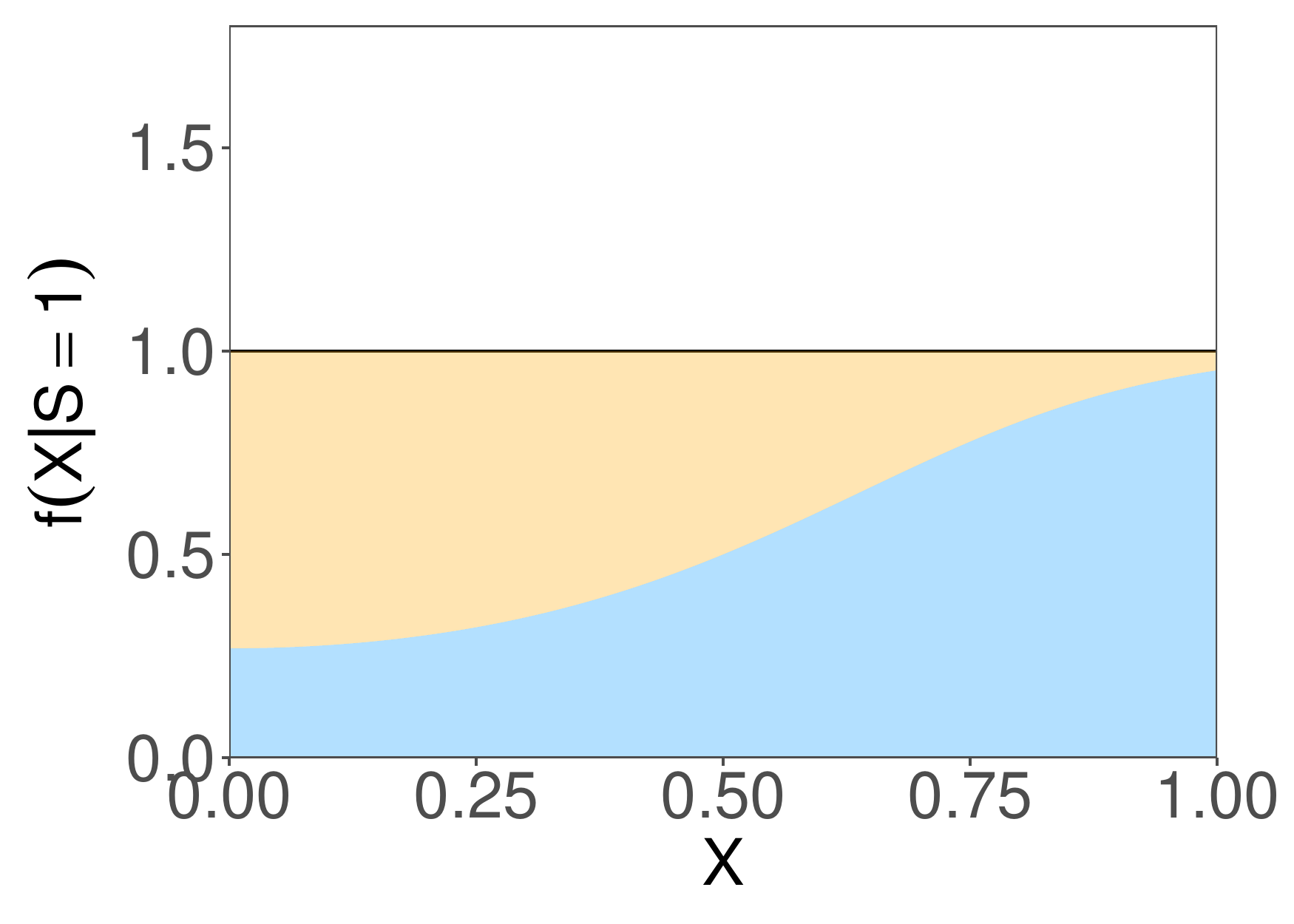}}%
      \quad
    \subfigure[$f(X\mid S=2)$]{\label{fig:den2}
      \includegraphics[width=0.3\linewidth]{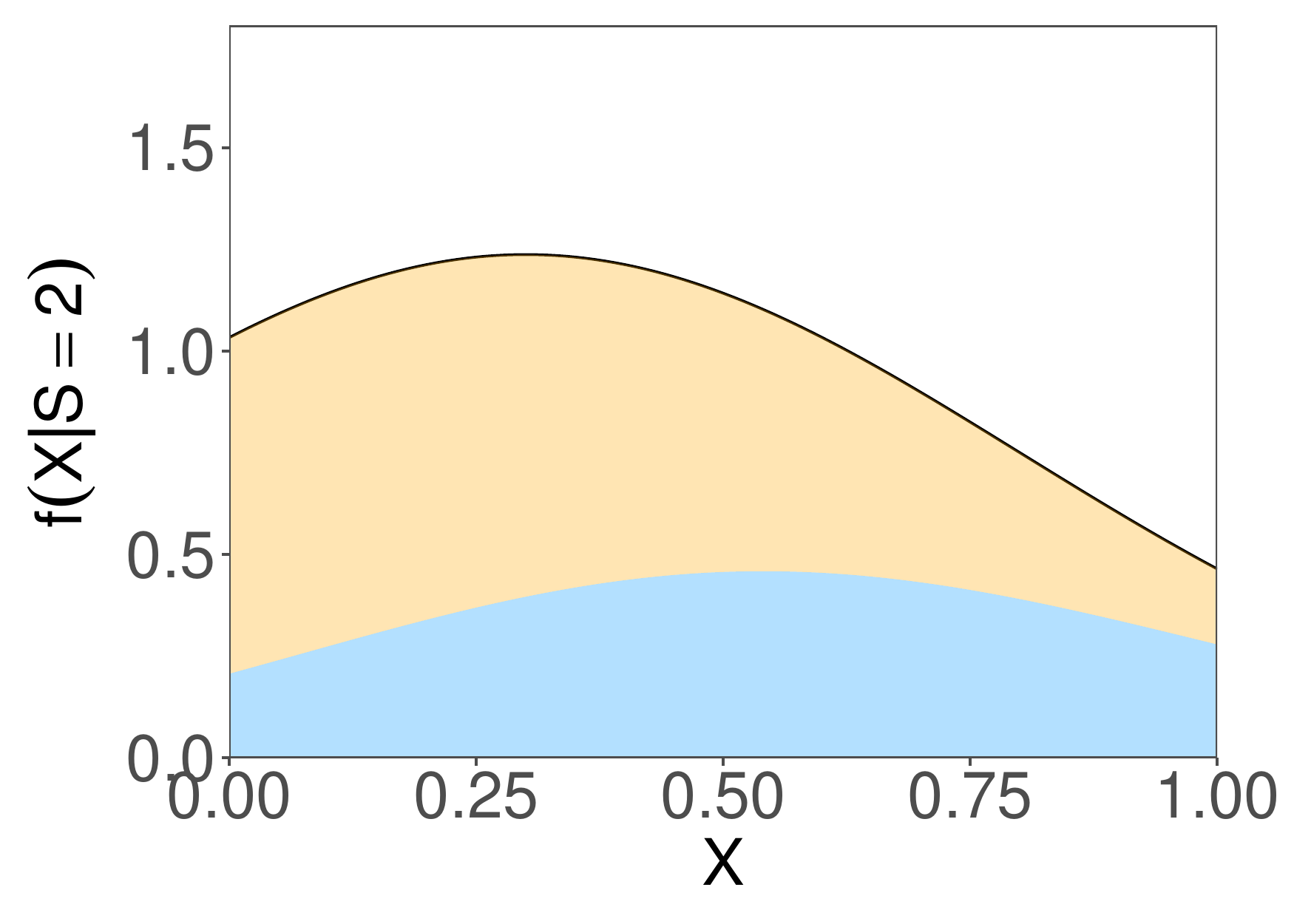}}\\
    \subfigure[$I_S(X)$]{\label{fig:info}
      \includegraphics[width=0.3\linewidth]{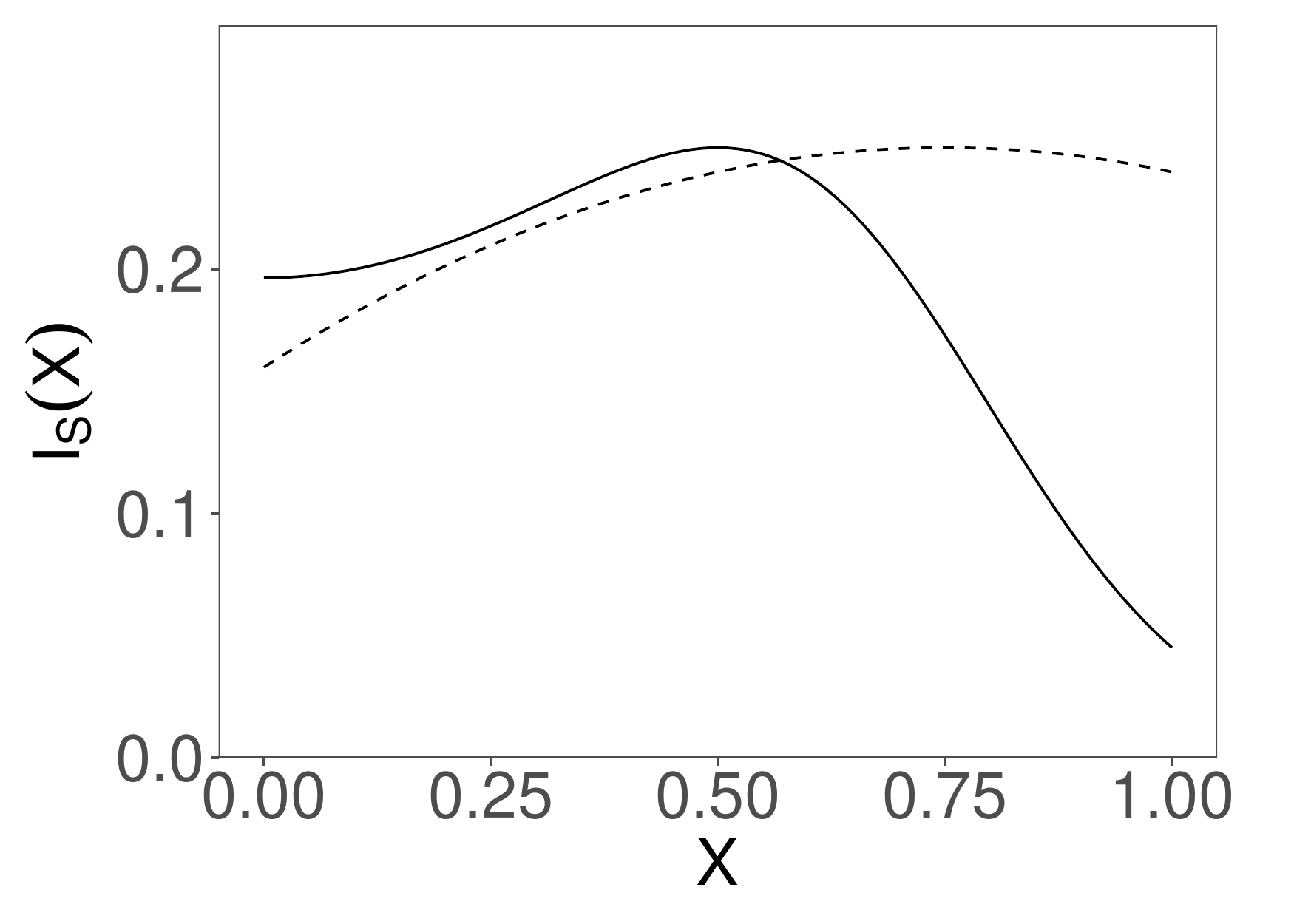}}
      \quad
          \subfigure[$R_S(X)$]{\label{fig:ratio}
      \includegraphics[width=0.3\linewidth]{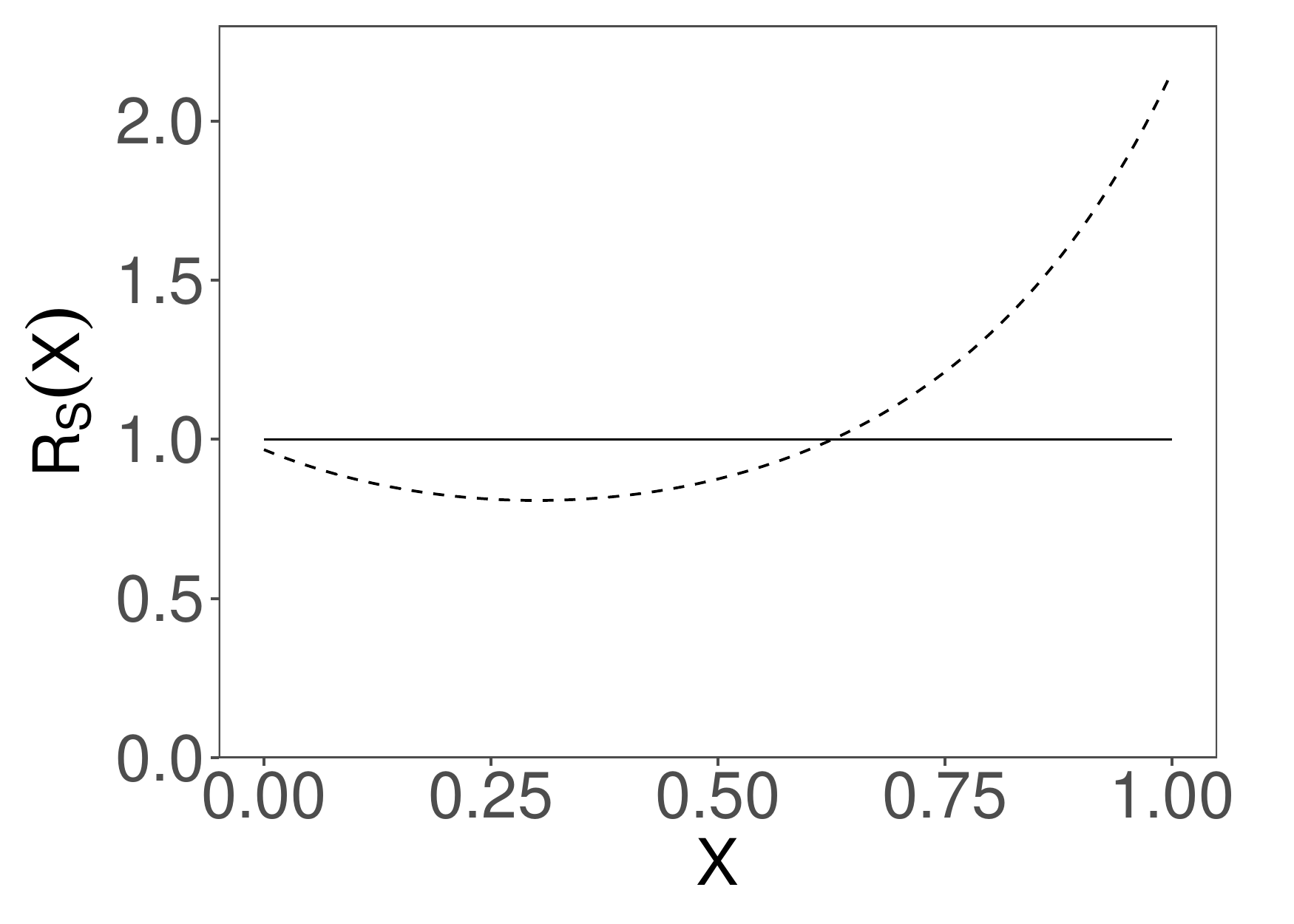}}
      \quad
    \subfigure[$w_S(X)$]{\label{fig:weight}
      \includegraphics[width=0.3\linewidth]{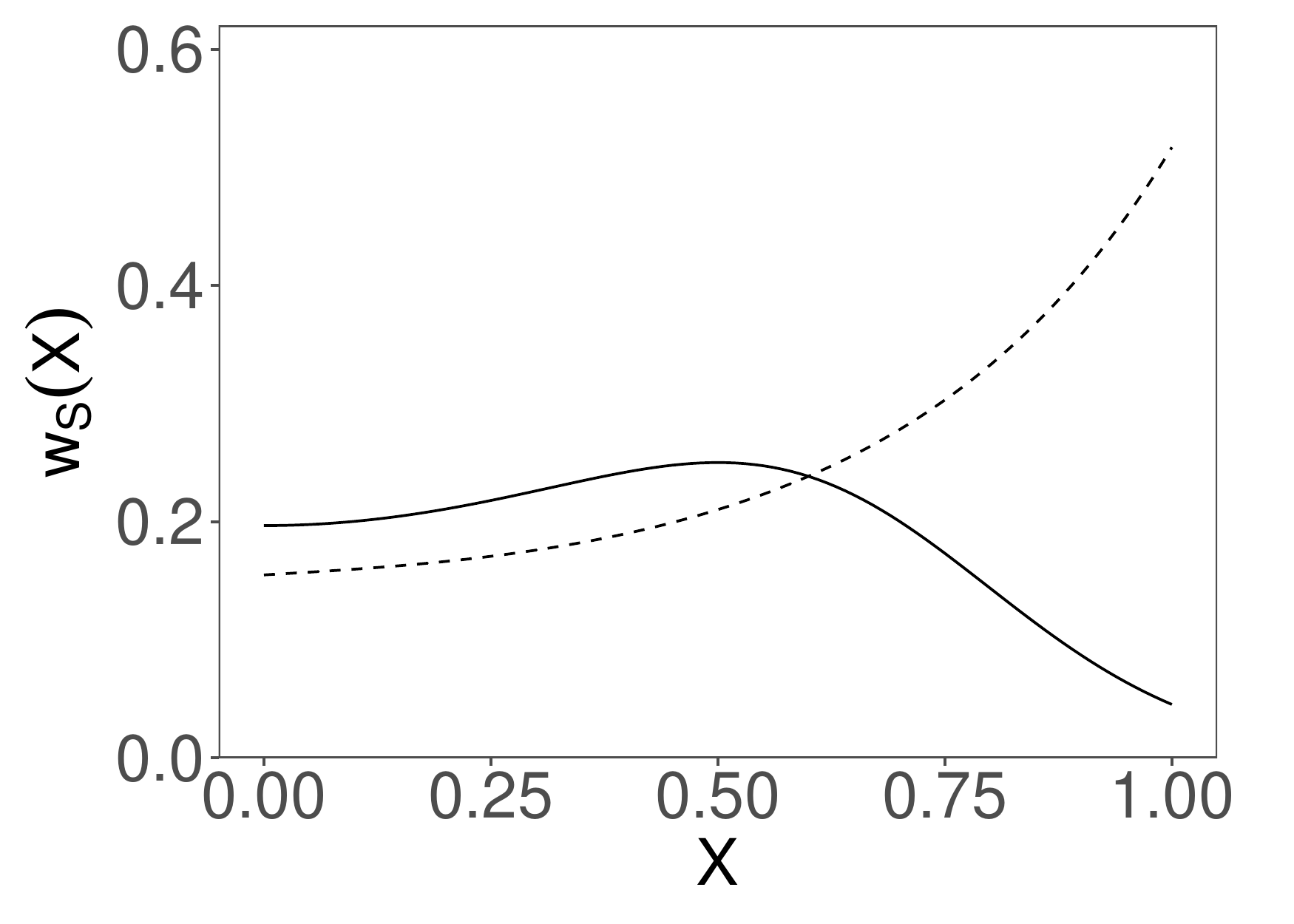}}
  }
\end{figure}

Our proposed weighting function $w_s(X)$ can be decomposed
into the product of two components: the transfer term $R_s(X)=f(X\mid S=1)/f(X\mid S=s)$ and  the information term  denoted by $I_s(X)$. The transfer term characterizes
the imbalance of populations between the $s$-th source and the target one. The information term
characterizes the amount of causal information satisfying that $I_{s_1}(\vec x)/I_{s_2}(\vec x)= \mathcal{B}_{s_2,\vec x}/\mathcal{B}_{s_1,\vec x}$, therefore assigns larger weights to the samples containing more causal information. To match with intuition, for observations  with covariates $X=\vec x$ in the $s$-th dataset, their  weights  are determined by:

\begin{itemize}
\item Density ratio corresponding to $f(\vec x\mid S=1)/f(\vec x\mid S=s)$.
This term enables us to pay more attention to the subpopulations that account for a larger share of the target population. 
Individuals that do not belong to the target population are discarded because their weights become zero.
Within the target dataset $S=1$, each observation is equally treated on this term.

\item Degree of imbalance in treatment assignment  corresponding to ${p}_{1\mid s}({\vec x})$. The weighting function prefers to reward a more balanced treatment  assignment mechanism. To illustrate, suppose that $V_{a\mid s}(\vec x)=1$ and then  the term $[1/{p}_{1\mid s}({\vec x})+1/\{1-{p}_{1\mid s}({\vec x})\}]^{-1}$ degenerates to ${p}_{1\mid s}({\vec x})\{1-{p}_{1\mid s}({\vec x})\}$, which is maximized at ${p}_{1\mid s}({\vec x})=0.5$.

\item Noise corresponding to $V_{1\mid s}(\vec x)$ and $V_{-1\mid s}(\vec x)$. The noise of potential outcomes varies across subpopulations and datasets, and our weighting function assigns larger  weights to those with less noise in a subtle way that cooperates with the information of treatment assignment.
\end{itemize}

\begin{remark}
When $K=1$ and $Z_1=\varnothing$, the setting degenerates to the classic problem of estimating CATE on a single dataset.  In this case, our proposed direct learning method can be regarded as a weighted version  of  \cite{meng2020doubly}, utilizing a weighting function $w(X)\propto[V_1(X)/p_1(X)+V_{-1}(X)/\{1-p_1(X)\}]^{-1}$. In the numerical experiments reported below, we show that the use of our proposed weight can significantly improve the performance of  direct learning.
\end{remark}

\subsection{Algorithm}

We propose a two-step procedure called Weighted Multi-source Direct Learner (WMDL) for  estimating  treatment effect functions from heterogeneous multi-source data, which is summarized in Algorithm \ref{algo:1}.
As a first step, we 
construct a pseudo-outcome and a weight value for each data point by plugging in the estimators of nuisance functions.  The nuisances are variation independent to CATE, including the main effect functions $m_s(X,Z_s)$, treatment propensity scores $p_{a\mid s}(X,Z_s)$ and weighting functions $w_s(X)$. 
By Equation \eqref{eq3}, estimating weighting function is equivalent to estimating selection propensity scores $\pi_s(X)$, conditional treatment propensity scores $p_{a\mid s}(X)$ and conditional outcome variances $V_{a\mid s}(X)$. Alternatively, one can also choose to model the conditional density ratios instead of the selection propensity scores.  The working models used to construct weighting functions only affect the efficiency, but not the consistency. To obtain nuisance estimators, we split each dataset into $G$ even folds, and   estimates are fit on data excluding the fold where the data point lies.
The data splitting technique is commonly used when learning nuisance estimators \citep{kallus2022robust}.

\begin{algorithm}
\caption{Weighted Multi-source Direct Learner}\label{algo:1}
\KwIn{Datasets $\mathcal{O}_{s},s=1,\cdots,K$,  nuisance function learners, regression learner}
\KwOut{$\hat\delta(X)$}
\For{$s=1,\cdots,K$}{
 Use dataset $\mathcal{O}_s$ to construct nuisance estimators $\hat{m}_s$, $\hat{p}_{a\mid s}$ and $\hat{w}_s$\; 
 
Pseudo-outcome $\widetilde{Y}_i=A_i\{Y_i-\hat{m}_s(X_i,Z_{s,i})\}$
        and $\widetilde{W}_i=\hat{w}_s(X_i)/\hat{p}_{A_i\mid s}(X_i,Z_{s,i})$\;
        
 Set $\mathcal{P}_s=\{\widetilde{Y}_i,\widetilde{W}_i,X_i\}_{i\in \mathcal{O}_s}$\;
}
Set $\mathcal{P}=\mathcal{P}_1 \cup \cdots \cup \mathcal{P}_K$\;

Fit regression learner on  $\set{P}$ to obtain  $\hat\delta(X)=\operatorname{argmin}_l\sum_{i\in \set{P}}\left[\widetilde{W}_i\left\{\widetilde{Y}_i-l(X_i)\right\}^2\right]$\;
\end{algorithm}

Then given a regression learner, for example, random forest, we  regress the pseudo-outcomes on the covariates $X$ by solving a weighted least square problem.
The regression learner is  specified independently of those nuisance function learners, which affords us significant generality.
The algorithm offers us a powerful tool for causal data fusion, enjoying the advantages of easy implementation, double robustness, and model flexibility.

\subsection{Direct Learning for Heterogeneous Causal Data Fusion}

In this part, we turn to investigate heterogeneous causal data fusion, where Assumption \ref{assp:3} does not hold anymore. We set the first data source $S=1$ as the target one, and hence the causal quantity of interest is $\delta_1(X)$.
We rewrite $\delta_S(X)$ as $\delta(X,S)$, then analogous to the \theoremref{thm:1},  under Assumptions \ref{assp:1} and \ref{assp:2}  one can obtain the function $\delta(X,S)$  by solving
\begin{equation*}
\mathop{\arg\min}_{l \in\{\mathcal{X}\times \mathcal{S} \rightarrow \mathbb{R}\}}
 {E}\left[\frac{w_{S}({X})}{{p}_{A\mid S}({X},Z_S)} \left\{\frac{Y-{m}_S({X},Z_S)}{A}- l({X},S)\right\}^{2}\right].
\end{equation*}

The inclusion of variable $S$ assists us in both distinguishing the heterogeneity of causal effects between different datasets and capturing the commonly shared structures.
Therefore, for heterogeneous causal data fusion,  we can still apply the proposed WMDL to estimate CATE in a direct, model-flexible, and robust way.
To be specific, the nuisance estimators remains the same as the homogeneous causal data fusion setting, and the estimation procedure for CATE is similar to Algorithm \ref{algo:1}, except that we turn to solve $\hat\delta(X,S)=\operatorname{argmin}_l\sum_{i\in \set{P}}[\widetilde{W}_i\{\widetilde{Y}_i-l(X_i,S_i)\}^2]$  in the last step. 
We take  $\hat\delta(x,1)$ as the resulted estimator of the treatment effect function $\delta_1(x)$; it  enjoys the property of double robustness  against misspecification of nuisance models corresponding to $\mathcal{O}_1$. In the numerical experiments, we demonstrate that WMDL can effectively integrate information from multiple data sources, resulting in more accurate estimates than a single data source.

\section{Experiments}

\subsection{Settings}
In this part, we conduct a broad simulation comparing the weighted multi-source direct learner  with other popular methods under both homogeneous and heterogeneous causal data fusions. 
We set the number of data sources $K=10$, 
and draw the same number of independent observations for each data source from the following data generating process:
$$
X \sim \operatorname{Unif}\left([-1,1]^{4}\right) \text{ for } S=1, \quad
X \sim N_{[-1,1]^{4}}(\mu_S,I) \text{ for } S=2,\dots,K,
$$
where $\mu_s \sim N(0,\sigma^2 I)$ with $\sigma=0.3$, $N_{[-1,1]^{4}}$ stands for the four-dimensional truncated normal distribution and $I$ denotes the identity matrix.  For each 
causal data fusion, we consider two scenarios: the Scenario \uppercase\expandafter{\romannumeral1} in which there are no
source-specific covariates with $Z_s=\varnothing$, and the Scenario \uppercase\expandafter{\romannumeral2}
 in which each data source has a  source-specific covariate $Z_s\sim N_{[-1,1]}(0, 1)$ for $s=1,\dots,K$.

Under both  homogeneous and heterogeneous causal data fusions, we generate the treatment $A$  in each dataset by $ \operatorname{Bernoulli}\left(\operatorname{expit}\{(X,Z_s)^{\operatorname{T}}\beta\}\right)$, where $\operatorname{expit}(\cdot) = \exp(\cdot)/\{1+ \exp(\cdot)\}$, and the parameter $\beta$ is randomly generated by the normal distribution $N(0,I)$. Then we generate the outcome $Y$ by \equationref{3.1}; the main effect function $m_s$'s detailed form is provided in the supplementary material, the random noise $\varepsilon_s\sim N(0, \sigma_s^2)$  with $\sigma_s=0.1$,  and the treatment effect functions
\begin{align*}
	\delta(X) =& (X_1+X_2+X3)\cdot \mathbbm{1}(X_1<0.5) + X_4,\\
	\delta_s(X) =& X_1\cdot \mathbbm{1}(X_1<0.5)\mathbbm{1}(\set{S}_1) + X_2\cdot \mathbbm{1}(X_2<0.5)\mathbbm{1}(s\leq 7)  \\
		&+ X_3\cdot \mathbbm{1}(X_3<0.5) + X_4\cdot \mathbbm{1}(\set{S}_1) + 2\cdot\mathbbm{1}(X_1<0)\mathbbm{1}(\set{S}_2),
\end{align*}
for homogeneous and  heterogeneous causal data fusions, respectively, 
where $\mathcal{S}_1 = \{s: s {\ \operatorname{mod} \ }2=1\}$ and $\mathcal{S}_2 = \{s: s {\ \operatorname{mod}\ }2=0\}$.


We demonstrate the numerical performance by comparing it with the following widely-used methods. The meta-learner methods decompose the CATE into several regression and classification problems solved by any supervised learning method, such as T-learner (TL), S-learner (SL), and X-learner (XL), see \cite{kunzel2019metalearners} for details.
Causal forest (CF) is another popular approach for estimating the CATE based on random forest \citep{wager2018estimation}.
For each method, we take the following two strategies to integrate datasets as the benchmark: (i) directly combine all the data and then apply the method to learn CATE, and (ii) include the source indicator as an additional predictor. We add ``$-s$'' at the end of the abbreviation (e.g., XL-s) to indicate the inclusion of  $S$. 
To assess  the role of proposed weighting function $w_S(X)$, we also take the multi-source direct learner using $w_S(X)=1$ in \algorithmref{algo:1} as a benchmark, 
referred to as MDL. 
Furthermore, to reveal the benefits of integrating multiple sources, we apply direct learners only on the target dataset, then MDL/WMDL degenerates to  the robust direct learning (DL) by \citet{meng2020doubly} and the weighted robust direct learning (WDL).
For the direct learners (WMDL, MDL, WDL and DL) and meta-learners (TL, SL, XL), we all use XGBoost for building the CATE and conditional mean models. 
We set hyperparameters for XGBoost as follows: the learning rate is 0.01, the maximum depth of a tree is 6  and the max number of boosting iterations is 20000.
We evaluate by replicating 100 times, each time on an independent test dataset of sample size $1000$ generated from the  target source. We use the mean square error (MSE), an empirical version of the $L_2$ distance, 
$${\operatorname{MSE}} = 
	n^{-1}\sum_{i=1}^n \left\{\hat\delta_1(X_i) - \delta_1(X_i)\right\}^2,
$$
as the performance metric.

\subsection{Results}

We summarize the average of mean squared error and the corresponding standard deviation of the above experiments in \tableref{tab:1}.

\begin{table}[h]
\centering
\caption{Mean and standard deviation of MSE under sample sizes 3000 and 5000}
\label{tab:1}
\begin{tabular}{lccccccccc}\toprule
& \multicolumn{4}{c}{Homogeneous} && \multicolumn{4}{c}{Heterogeneous}  \\ 
  \midrule
& \multicolumn{2}{c}{\uppercase\expandafter{\romannumeral1}} & \multicolumn{2}{c}{\uppercase\expandafter{\romannumeral2}} &&
\multicolumn{2}{c}{\uppercase\expandafter{\romannumeral1}} &
\multicolumn{2}{c}{\uppercase\expandafter{\romannumeral2}}\\ 
  \cmidrule{2-5}  \cmidrule{7-10} 
 & 3000 & 5000 & 3000 & 5000 && 3000 & 5000 & 3000 & 5000 \\ 
  \cmidrule{2-5}  \cmidrule{7-10} 
WMDL & \bf{0.084} & \bf{0.045} & \bf{0.106} & \bf{0.064} && \bf{0.163} & \bf{0.102} & \bf{0.289} & \bf{0.192} \\ 
& (0.008) & (0.003) & (0.009) & (0.006) && (0.035) & (0.014) & (0.066) & (0.030) \\ 
MDL & 0.104 & 0.061 & 0.143 & 0.089 && 0.198 & 0.121 & 0.380 & 0.251 \\ 
& (0.009) & (0.005) & (0.013) & (0.009) && (0.037) & (0.016) & (0.084) & (0.045) \\ 
WDL & 0.227 & 0.148 & 0.458 & 0.303 && 0.226 & 0.143 & 0.451 & 0.303 \\ 
& (0.060) & (0.026) & (0.096) & (0.051) && (0.053) & (0.020) & (0.089) & (0.049) \\ 
DL & 0.275 & 0.175 & 0.532 & 0.346 && 0.275 & 0.166 & 0.532 & 0.350 \\ 
& (0.064) & (0.030) & (0.112) & (0.057) && (0.058) & (0.023) & (0.094) & (0.059) \\ 
CF & 0.171 & 0.159 & 0.198 & 0.155 && 0.959 & 0.948 & 0.819 & 0.780 \\ 
& (0.011) & (0.009) & (0.014) & (0.010) && (0.038) & (0.039) & (0.043) & (0.038) \\ 
XL & 0.394 & 0.356 & 0.478 & 0.359 && 1.225 & 1.141 & 1.136 & 0.988 \\ 
& (0.024) & (0.029) & (0.039) & (0.026) && (0.049) & (0.056) & (0.077) & (0.054) \\ 
TL & 0.591 & 0.500 & 0.781 & 0.593 && 1.448 & 1.331 & 1.488 & 1.264 \\ 
& (0.039) & (0.037) & (0.060) & (0.039) && (0.071) & (0.066) & (0.093) & (0.067) \\ 
SL & 0.267 & 0.226 & 0.248 & 0.202 && 1.058 & 1.005 & 0.893 & 0.831 \\ 
& (0.028) & (0.023) & (0.023) & (0.020) && (0.044) & (0.047) & (0.055) & (0.042) \\ 

CF-s & 0.137 & 0.105 & 0.160 & 0.129 && 0.637 & 0.474 & 0.625 & 0.425 \\ 
& (0.009) & (0.006) & (0.010) & (0.009) && (0.067) & (0.047) & (0.102) & (0.042) \\ 
XL-s & 0.119 & 0.080 & 0.708 & 0.649 && 0.213 & 0.131 & 0.858 & 0.612 \\ 
& (0.028) & (0.016) & (0.215) & (0.168) && (0.038) & (0.020) & (0.280) & (0.191) \\ 
TL-s & 0.285 & 0.215 & 1.237 & 1.066 && 0.350 & 0.231 & 1.372 & 1.049 \\ 
& (0.074) & (0.047) & (0.284) & (0.209) && (0.071) & (0.042) & (0.343) & (0.195) \\ 
SL-s & 0.158 & 0.155 & 0.438 & 0.464 && 0.420 & 0.355 & 0.560 & 0.503 \\ 
& (0.019) & (0.016) & (0.194) & (0.155) && (0.062) & (0.047) & (0.170) & (0.134) \\ 
\bottomrule
\end{tabular}
\end{table}

From \tableref{tab:1}, we reach the following conclusions:  
\begin{itemize}
    \item Compared with the benchmarks, the proposed WMDL results in the smallest mean squared error on average in all scenarios, demonstrating its advantages  in terms of  precision under both homogeneous and heterogeneous causal data fusions. Also, MDL and WMDL typically have a smaller standard deviation than other methods, which implies that multi-source direct learning may lead to a more stable performance. 

\item By comparing WMDL/WDL to MDL/DL, as expected, the use of our proposed information-aware weighting function significantly improves both the accuracy and stability of the CATE estimates, regardless of whether multiple sources or a single source is analyzed. 

\item The multi-source approach (MDL/WMDL) outperforms the single-source approach (DL/WDL) substantially, where the MSE decreases considerably in homogeneous cases and reduces to a certain degree in heterogeneous cases. This indicates that our proposed method can effectively integrate causal information from multiple sources, and may lead to efficiency gains even when the CATE differs across data sources.

\item The inclusion of the source indicator as a covariate in CF, TL, SL and XL can generally promote the performance in the Scenario \uppercase\expandafter{\romannumeral1}, but may lead to larger MSEs in the Scenario \uppercase\expandafter{\romannumeral2}. 
In contrast, the WMDL behaves well across scenarios. It suggests that our method can provide a  simple but powerful way to make effective use of source-specific covariates, which only requires these covariates when modeling the nuisance functions but  not the CATE.
\end{itemize}


\figureref{fig:plots} presents the MSE of MDL and WMDL with increasing sample size. The WMDL outperforms the MDL under both homogeneous and heterogeneous settings, which highlights the effectiveness of
our proposed causal information-aware function. The box plots in \figureref{fig:plots} also show the variability of MD-learning can be reduced with the weighting function. These numerical results show the potential application of efficiency theory in \sectionref{sec:3.4} to CATE problems.

\begin{figure}[htp]
\floatconts
  {fig:plots}
  {\caption{Boxplots of mean square error for multi-source direct learning with and without the causal information-aware weighting function.}}
  {%
    \subfigure[Homogeneous causal data fusion]{\label{fig:image-a}%
      \includegraphics[width=0.45\linewidth]{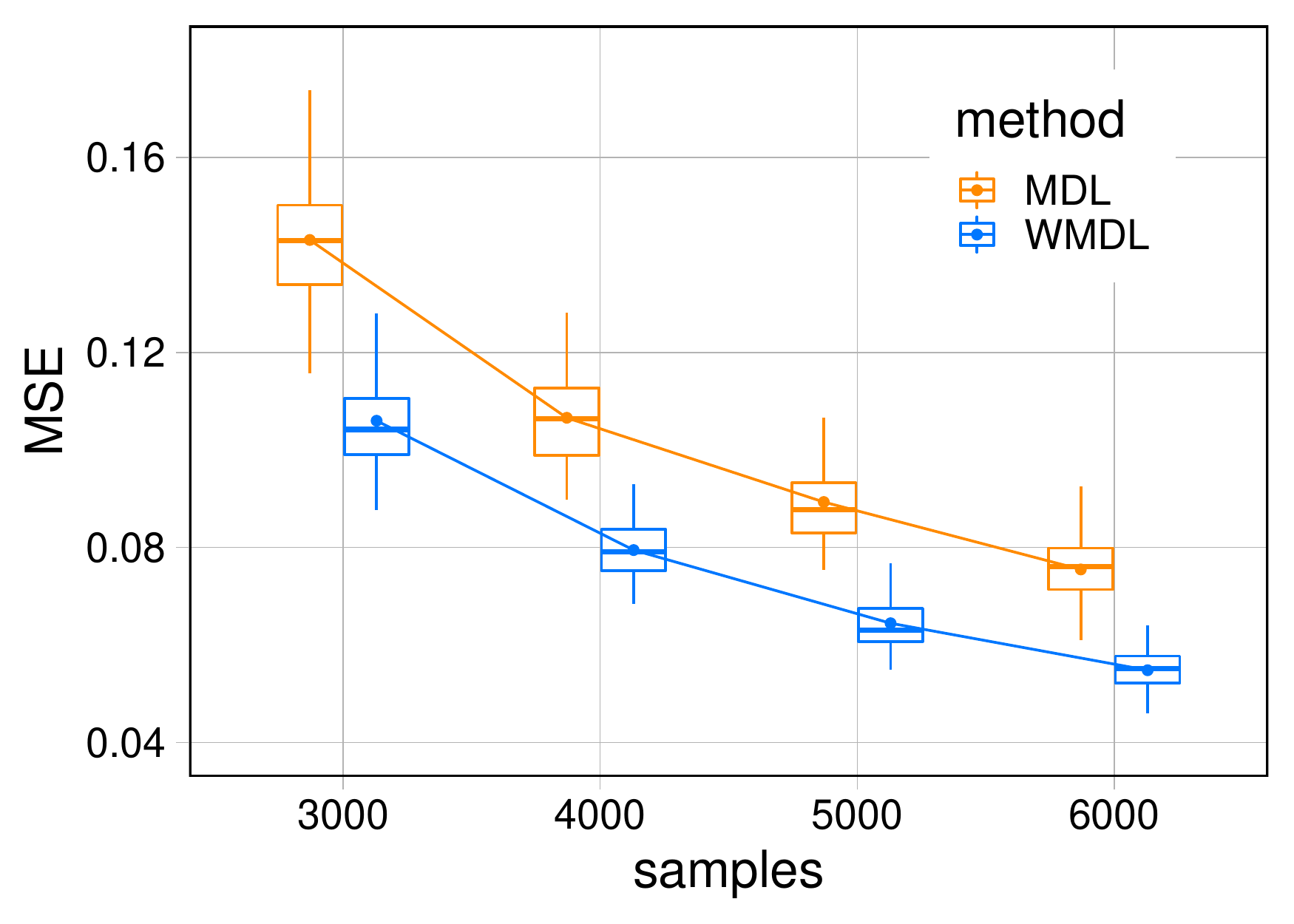}}%
    \qquad
    \subfigure[Heterogeneous causal data fusion]{\label{fig:image-b}%
      \includegraphics[width=0.45\linewidth]{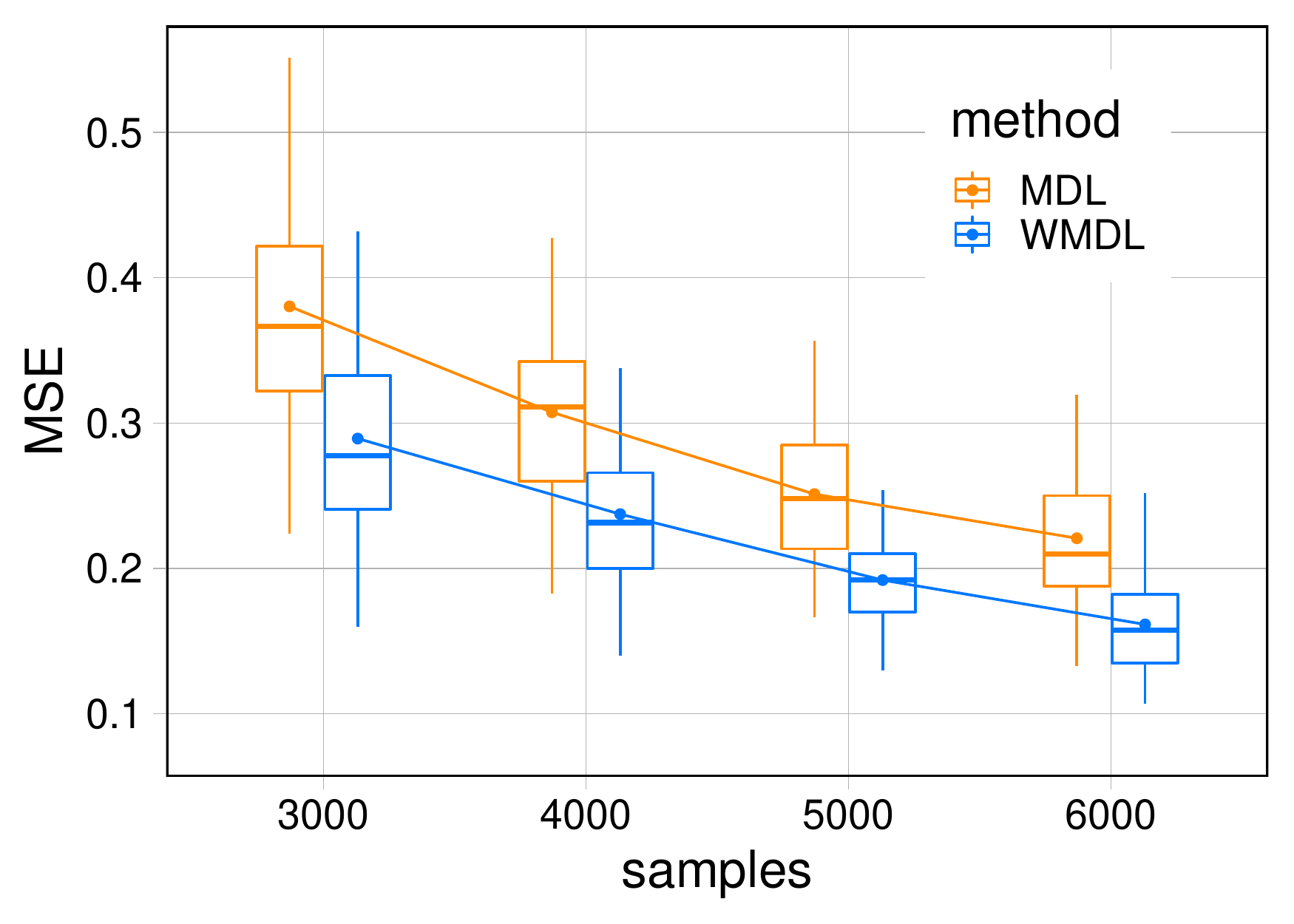}}
  }
\end{figure}

\section{Discussion}

We provide  a simple yet powerful algorithm, the WMDL,  for causal data fusion. 
In this paper, the target dataset contains the outcome data  under different treatments. However, in some cases, only the covariates of the target population are available \citep{dahabreh2020towards}. Therefore transferring the causal inference from other sources to the target population is also of interest. We highlight that our proposed WMDL can also be applied to \textit{causal transfer learning}, see details in Appendix \ref{apd:first}.
Our approach is double-weighted, with treatment propensity scores adjusting for bias and enhancing robustness, and causal information-aware weighting functions improving efficiency.
In the future, we may consider adding the uncertainty of  models to  weighting functions to achieve more stable performance. 

\section*{Acknowledgements}
This work was supported by Ant Group through Ant Research Intern Program.  We thank the editors and three anonymous referees for their constructive suggestions and   comments.
The proof of Theorem \ref{thm:1} and more simulation details are available in the supplementary material.

\bibliography{acml22}

\appendix

\section{Discussions on Causal Transfer Learning}\label{apd:first}

\setcounter{theorem}{0}
\renewcommand{\thetheorem}{A.\arabic{theorem}}   
\setcounter{equation}{0}
\renewcommand{\theequation}{A.\arabic{equation}}   

Suppose we have individual-level
data under different treatments from multiple sources  as well as covariate data from  a target population.
We aim to synthesize findings across multiple observational datasets and transport causal inferences to the target population.
Such scenarios are common in real applications,   see \cite{dahabreh2020towards}. 
In this section, we provide a detailed discussion on how to apply our proposed framework to transfer causal inference. 

To formalize the problem, we keep the notation consistent with the main text and let the source indicator $S=0$ to indicate the target population.
Individual-level data under different treatments are collected in $K$ mutually independent trial datasets, but for the target population, only covariate information is available.
We summarize the observed datasets as $\mathcal{O}_0 = \{ X_i, i = 1, \cdots, n_0\}$ and $\mathcal{O}_s = \{(Y_i,A_i, X_i, Z_{s,i}), i = 1, \cdots, n_s\}$ for $s= 1,\cdots,K$. 
Then the joint distribution of observed data is $f(X,S=0)^{\mathbbm{1}(S=0)} \prod_{s=1}^{K}\left[f(Y,A,X, Z_s,S=s)\right]^{\mathbbm{1}(S=s)}$.
Our parameter of
interest is the CATE function on the target population $\tau_0(X) = E\{Y(1) - Y(-1) | X,S = 0\}$. To identify $\tau_0(X)$, we impose the following assumptions.

\begin{assumption}[Ignorability]\label{assp:a1}
$Y(a)\perp A\mid X,Z_s,S=s$ for each $s\in\{1,\cdots,K\}$.
\end{assumption}
\begin{assumption}[Treatment Positivity]\label{assp:a2}
 $P(A=a \mid X=\vec x,Z_s,S=s)>0$ for all $\vec x$ such that $f(X=\vec x \mid S=s)f(X=\vec x \mid S=0)>0$, where $a\in\{1,-1\}$ and $s\in\{1,\cdots,K\}$.
\end{assumption}
\begin{assumption}[Population Positivity]\label{assp:a3}
$\sum_{s=1}^{K}f(X=\vec x \mid S=s)>0$ for all $\vec x$ such that $f(X=\vec x \mid S=0)>0$.
\end{assumption}

\begin{assumption}[Homogeneity of CATE]\label{assp:a4}
$\delta_0(X)=\cdots=\delta_K(X)$ almost surely.
\end{assumption}

Notably, in addition to the assumptions required in homogeneous causal data fusion, we impose the \textit{population positivity} assumption, which implies that similar individuals exist in at least one trial dataset for those in the target population.
Under Assumptions \ref{assp:a1}--\ref{assp:a4}, the CATE function $\tau_0(X)$ is identified.

The estimation procedures are the same as in Algorithm \ref{algo:1} of the main text, except that the information-aware weighting function
\begin{align}
w_s(X)&\propto \frac{P(S=s)}{
P(S=0)}\frac{\pi_0(X)}{\pi_s(X)} \left\{\frac{V_{1\mid s}(X)}{{p}_{1\mid s}({X})}+\frac{V_{-1\mid s}(X)}{{p}_{-1\mid s}({X})}  \right\}^{-1} \label{eq.a1}\\
&= \frac{f(X\mid S=0)}{f(X\mid S=s)} \left\{\frac{V_{1\mid s}(X)}{{p}_{1\mid s}({X})}+\frac{V_{-1\mid s}(X)}{{p}_{-1\mid s}({X})}  \right\}^{-1}\label{eq.a2}.
\end{align}
The density ratio ${f(X\mid S=0)}/{f(X\mid S=s)}$ enables us to pay more attention to the subpopulations that account for a larger share of the target population. We assign a weight of zero to individuals not belonging to the target population, i.e., those with covariate $\vec x$ such that $f(X=\vec x\mid S=0)=0$.
Analogous to the causal data fusion, the specification of $w_s(X)$ only relates to the efficiency,  but not the consistency. 
One can either model  propensity scores and then estimate $w_s(X)$  via Equation \eqref{eq.a1}, or model  densities and then estimate $w_s(X)$   via Equation \eqref{eq.a2}. Under Assumptions \ref{assp:a1}--\ref{assp:a4}, the obtained $\hat \delta(X)$ shares the same properties as those described in the main text.

\end{document}